\documentclass{article}

\usepackage[final, nonatbib]{neurips_2021}

\newcommand*\samethanks[1][\value{footnote}]{\footnotemark[#1]}

\usepackage{amsmath, amsthm, amssymb}

\usepackage{expl3}  %
\ExplSyntaxOn
  _to_str:N
\ExplSyntaxOff
\usepackage{ifthen}
\usepackage{pgffor}

\usepackage[acronym,smallcaps,nowarn,section,nonumberlist]{glossaries}
\glsdisablehyper  %
\makeglossaries

\setacronymstyle{long-short}

\newacronym{auc}{AUC}{Area Under the Curve}

\newacronym{bert}{BERT}{Bidirectional Encoder Representations from Transformers}
\newacronym{bg}{BG}{block groups}
\newacronym{bow}{BOW}{Bag of Words}
\newacronym{boe}{BOE}{Bag of Embeddings}
\newacronym{borep}{BOREP}{Bag of Random Embedding Projections}
\newacronym{bn}{BN}{Batch Normalization}
\newacronym{byol}{BYOL}{Bootstrap Your Own Latent}

\newacronym{cdf}{CDF}{Cumulative Distribution Function}
\newacronym{cka}{CKA}{Centered Kernel Alignment}
\newacronym{cnn}{CNN}{convolutional neural network}
\newacronym{cvae}{CVAE}{Conditional Variational Auto-Encoder}

\newacronym{dag}{DAG}{Directed Acyclic Graph}
\newacronym{dino}{DINO}{DIstillation with NO labels}
\newacronym{dgm}{DGM}{Directed Graphical Model}
\newacronym{dnn}{DNN}{Deep Neural Network}

\newacronym{ecfp4}{ECFP4}{Extended Connectivity Fingerprint}
\newacronym{ehr}{EHR}{Electronic Health Record}
\newacronym{esn}{ESN}{Echo State Network}

\newacronym{fft}{FFT}{Fast Fourier Transform}
\newacronym{fs}{FS}{FastSent}
\newacronym{ft}{FT}{Fourier Transform}

\newacronym{gat}{GAT}{Graph Attention Network}
\newacronym{gcn}{GCN}{Graph Convolutional Network}
\newacronym{gdl}{GDL}{Geometric Deep Learning}
\newacronym{ggnn}{GGNN}{Gated Graph Neural Network}
\newacronym{glove}{GloVe}{Global Vectors for Word Representation}
\newacronym{gnn}{GNN}{Graph Neural Network}
\newacronym{gru}{GRU}{Gated Recurrent Unit}

\newacronym{hmm}{HMM}{Hidden Markov Model}
\newacronym{hsic}{HSIC}{Hilbert-Schmidt Independence Criterion}

\newacronym{idf}{IDF}{Inverse Document Frequency}
\newacronym{ig}{IG}{Information Gain}
\newacronym{isan}{ISAN}{Input Switched Affine Network}

\newacronym{jsd}{JSD}{Jensen-Shannon Divergence}

\newacronym{kld}{KLD}{Kullback-Leibler Divergence}
\newacronym{knn}{KNN}{$K$-Nearest Neighbour}

\newacronym{lhs}{L.H.S.}{left hand side}
\newacronym{lstm}{LSTM}{Long Short Term Memory Network}
\newacronym{lv}{LV}{latent variable}

\newacronym{map}{MAP}{Maximum a Posteriori}
\newacronym{mc}{MC}{Monte Carlo}
\newacronym{mdl}{MDL}{Minimum Description Length}
\newacronym{ml}{ML}{Machine Learning}
\newacronym{mlp}{MLP}{Multi-Layer Perceptron}
\newacronym{mle}{MLE}{Maximum Likelihood Estimation}
\newacronym{mnist}{MNIST}{Modified National Institute of Standards and Technology}

\newacronym{nlp}{NLP}{Natural Language Processing}
\newacronym{nn}{NN}{Neural Network}
\newacronym{nll}{NLL}{Negative Log Likelihood}
\newacronym{nce}{NCE}{Noise Contrastive Estimation}

\newacronym{ode}{ODE}{Ordinary Differential Equation}
\newacronym{oh}{OH}{One-Hot}

\newacronym{pdf}{PDF}{Probability Density Function}
\newacronym{pgm}{PGM}{Probabilistic Graphical Model}

\newacronym{relu}{ReLU}{Rectified Linear Unit}
\newacronym{rdf}{RDF}{Resource Description Framework}
\newacronym{rgb}{RGB}{Red Green Blue}
\newacronym{rgc}{RGC}{Relational Graph Convolution}
\newacronym{rgat}{RGAT}{Relational Graph Attention Network}
\newacronym{rgcn}{RGCN}{Relational Graph Convolutional Network}
\newacronym{rhs}{R.H.S.}{Right Hand Side}
\newacronym{rnn}{RNN}{Recurrent Neural Network}
\newacronym{roc}{ROC}{Receiver Operating Characteristic}
\newacronym{rv}{RV}{Random Variable}

\newacronym{sgd}{SGD}{Stochastic Gradient Descent}
\newacronym{sg}{SG}{SkipGram}
\newacronym{simclr}{SimCLR}{A Simple framework for Contrastive Learning of visual Representations}
\newacronym{ssl}{SSL}{self-supervised learning}
\newacronym{sl}{SL}{supervised learning}

\newacronym{st}{ST}{SkipThought}
\newacronym{sts}{STS}{Semantic Textual Similarity}
\newacronym{scon}{StochCon}{Stochastic Contrastive Learning}
\newacronym{supcon}{SupCon}{Supervised Contrastive Learning}

\newacronym{termfreq}{TF}{Term Frequency}
\newacronym{tfidf}{TF-IDF}{Term Frequency-Inverse Document Frequency}
\newacronym[plural=TPPs, descriptionplural={Temporal Point Processes}]{tpp}{TPP}{Temporal Point Process}

\newacronym{vae}{VAE}{Variational Auto-Encoder}

\newacronym{wirgat}{WIRGAT}{Within-Relation Graph Attention}
\newacronym{wl}{WL}{Weisfeiler-Lehman}

\usepackage{bm}
\usepackage[T1]{fontenc}  %
\usepackage[type1]{libertine}
\usepackage[libertine]{newtxmath}
\usepackage{mathtools}

\usepackage{algorithm}
\usepackage{algpseudocode}

\usepackage{pythonhighlight}

\DeclareRobustCommand{\mathup}[1]{\begingroup\changegreek\mathrm{#1}\endgroup}
\DeclareRobustCommand{\mathbfup}[1]{\begingroup\changegreekbf\mathbf{#1}\endgroup}
\DeclareRobustCommand{\mathbit}[1]{\bm{\mathit{#1}}}

\DeclareMathAlphabet{\mathsfit}{\encodingdefault}{\sfdefault}{m}{sl}
\SetMathAlphabet{\mathsfit}{bold}{\encodingdefault}{\sfdefault}{bx}{n}
\newcommand{\tens}[1]{\bm{\mathsfit{#1}}}

\newcommand{\constantvector}{\bm}               %

\newcommand{\constantmatrix}{\bm}               %
\newcommand{\constantmatrixgreek}{\mathbit}

\newcommand{\randomscalar}{\textnormal}         %
\newcommand{\randomscalargreek}{\mathup}

\newcommand{\randomvector}{\mathbf}             %
\newcommand{\randomvectorgreek}{\mathbfup}

\newcommand{\randommatrix}{\mathbf}             %
\newcommand{\randommatrixgreek}{\mathbfup}

\newcommand{\graphstyle}{\mathcal}              %

\newcommand{\tensorstyle}{\tens}                %

\newcommand{\setstyle}{\mathbb}                %

\def\alphabet{a,b,c,d,e,f,g,h,i,j,k,l,m,n,o,p,q,r,s,t,u,v,w,x,y,z}
\def\Alphabet{A,B,C,D,E,F,G,H,I,J,K,L,M,M,O,P,Q,R,S,T,U,V,W,X,Y,Z}
\def\greekalphabet{alpha,beta,gamma,delta,epsilon,varepsilon,zeta,eta,theta,vartheta,iota,kappa,varkappa,lambda,mu,nu,xi,pi,varpi,rho,varrho,sigma,varsigma,tau,upsilon,phi,varphi,chi,psi,omega}
\def\GreekAlphabet{Gamma,Delta,Theta,Lambda,Xi,Pi,Sigma,Upsilon,Phi,Psi,Omega}

\makeatletter
\def\changegreek{\@for\next:=\greekalphabet
	\do{\expandafter\let\csname\next\expandafter\endcsname\csname\next up\endcsname}}
\def\changegreekbf{\@for\next:=\greekalphabet
	\do{\expandafter\def\csname\next\expandafter\endcsname\expandafter{%
			\expandafter\bm\expandafter{\csname\next up\endcsname}}}}
\makeatother

\foreach \x in \alphabet {
	\expandafter\xdef\csname v\x\endcsname{\noexpand\ensuremath{\noexpand\constantvector{\x}}}

	\expandafter\xdef\csname ev\x\endcsname{\noexpand\ensuremath{\noexpand\x}}

	\ifthenelse{\equal{\x}{m}}{}{
		\expandafter\xdef\csname r\x\endcsname{\noexpand\ensuremath{\noexpand\randomscalar{\x}}}
	}

	\expandafter\xdef\csname rv\x\endcsname{\noexpand\ensuremath{\noexpand\randomvector{\x}}}
}

\foreach \x in \greekalphabet {
	\expandafter\xdef\csname v\x\endcsname{\noexpand\ensuremath{\noexpand\constantvector{\csname \x\endcsname}}}

	\expandafter\xdef\csname ev\x\endcsname{\noexpand\ensuremath{\noexpand{\csname \x \endcsname}}}

	\expandafter\xdef\csname r\x\endcsname{\noexpand\ensuremath{\noexpand\randomscalargreek{\csname \x\endcsname}}}

	\expandafter\xdef\csname rv\x\endcsname{\noexpand\ensuremath{\noexpand\randomvectorgreek{\csname \x\endcsname}}}
}

\def\vone{{\constantvector{1}}}

\foreach \x in \Alphabet {
	\expandafter\xdef\csname m\x\endcsname{\noexpand\ensuremath{\noexpand\constantmatrix{\x}}}

	\expandafter\xdef\csname em\x\endcsname{\noexpand\ensuremath{\noexpand\x}}

	\expandafter\xdef\csname rm\x\endcsname{\noexpand\ensuremath{\noexpand\randommatrix{\x}}}

	\expandafter\xdef\csname t\x\endcsname{\noexpand\ensuremath{\noexpand\tensorstyle{\x}}}

	\expandafter\xdef\csname g\x\endcsname{\noexpand\ensuremath{\noexpand\graphstyle{\x}}}

	\ifthenelse{\equal{\x}{E}}{}{
		\expandafter\xdef\csname s\x\endcsname{\noexpand\ensuremath{\noexpand\setstyle{\x}}}
	}

}

\foreach \x in \GreekAlphabet {
	\expandafter\xdef\csname m\x\endcsname{\noexpand\ensuremath{\noexpand\constantmatrixgreek{\csname \x\endcsname}}}

	\expandafter\xdef\csname rm\x\endcsname{\noexpand\ensuremath{\noexpand\randommatrixgreek{\csname \x\endcsname}}}
}

\DeclareMathOperator{\Tr}{Tr}

\DeclareRobustCommand{\pnorm}[2]{\ensuremath{\left\lVert#1\right\rVert}_{#2}}

\newcommand{\R}{\mathbb{R}}

\newcommand*{\tran}{^{\mkern-1.5mu\mathsf{T}}}

\DeclareMathOperator{\cka}{CKA}
\DeclareMathOperator{\hsic}{HSIC}
\DeclareMathOperator{\vecc}{vec}

\usepackage{float}
\usepackage[utf8]{inputenc} %
\usepackage[T1]{fontenc}    %
\usepackage[hidelinks]{hyperref} %
\usepackage{url}            %
\usepackage{booktabs}       %
\usepackage{amsfonts}       %
\usepackage{nicefrac}       %
\usepackage{microtype}      %
\usepackage{xcolor}         %

\usepackage{cleveref}

\usepackage[square, numbers]{natbib} %
\usepackage[]{graphicx}

\usepackage{todonotes}

\usepackage{comment}
\usepackage{subcaption}

\title{Do Self-Supervised and Supervised Methods Learn Similar Visual Representations?}

\author{%
  Tom George Grigg \thanks{Use footnote for providing further information
    about author (webpage, alternative address)---\emph{not} for acknowledging
    funding agencies.} \\
  \And
  Dan Busbridge \\
  \And
  Jason Ramapuram \\
  \And
  Russ Webb \\
   Apple\\ 
  \texttt{\{tgrigg, dbusbridge\}@apple.com} 
}

\author{%
  Tom George Grigg\,\thanks{Work completed during an internship at Apple.}\;\,\thanks{Equal contribution.
  Correspondence to 
  \href{tomgeorgegrigg@gmail.com}{tomgeorgegrigg@gmail.com}
  and
  \href{dbusbridge@apple.com}{dbusbridge@apple.com}.}
  \quad Dan Busbridge\samethanks
  \quad Jason Ramapuram
  \quad Russ Webb \\\\
  Apple\\
}

\begin{document}
\maketitle

\begin{abstract} 
Despite the success of a number of recent techniques for visual self-supervised deep learning, there has been limited investigation into the representations that are ultimately learned. 
By leveraging recent advances in the comparison of neural representations, we explore in this direction by comparing a contrastive self-supervised algorithm to supervision for simple image data in a common architecture. 
We find that the methods learn similar intermediate representations through dissimilar means, and that the representations diverge rapidly in the final few layers.
We investigate this divergence, finding that these layers strongly fit to their distinct learning objectives. We also find that the contrastive objective implicitly fits the supervised objective in intermediate layers, but that the reverse is not true.
Our work particularly highlights the importance of the learned intermediate representations, and raises critical questions for auxiliary task design.

\end{abstract}

\section{Introduction}

In the last two decades, progress in deep learning for visual tasks has primarily been driven by training convolutional neural networks (CNNs)  \citep{DBLP:conf/cvpr/DengDSLL009,DBLP:conf/iclr/DosovitskiyB0WZ21,DBLP:conf/cvpr/HeZRS16,DBLP:conf/nips/KrizhevskySH12,DBLP:journals/neco/LeCunBDHHHJ89,DBLP:journals/corr/RussakovskyDSKSMHKKBBF14, DBLP:conf/eccv/ZeilerF14} on large labelled datasets via \gls{sl}. 
More recently, \gls{ssl} algorithms have started to close the performance gap \citep{DBLP:conf/nips/AlayracRSARFSDZ20, DBLP:conf/nips/CaronMMGBJ20, DBLP:journals/corr/abs-2104-14294, DBLP:conf/icml/ChenK0H20,DBLP:conf/nips/ChenKSNH20, DBLP:conf/nips/GrillSATRBDPGAP20,DBLP:journals/corr/abs-2103-03230}. 
The success of these visual \gls{ssl}  algorithms raises important questions from a representation learning perspective: how are \gls{ssl}  methods building competitive respresentations without access to class labels? Do learned representations differ between \gls{sl} and \gls{ssl}? If so, can/should we encourage them to be similar? Do different \gls{ssl} objectives learn qualitatively different representations? In this work, we begin to shed light in this direction by comparing the representations of CIFAR-10 (C10) induced in a ResNet-50 (R50) architecture by \gls{sl} against those induced by SimCLR, a prominent contrastive \gls{ssl} algorithm. We find that:

\begin{itemize}
\item Post-residual representations are similar across methods, however residual (block-interior) representations are dissimilar; similar structure is recovered by solving different problems. 
\item Initial residual layer representations are similar, indicating a shared set of primitives.
\item The methods strongly fit to their distinct objectives in the final few layers, where SimCLR learns augmentation invariance and \gls{sl} fits to the class structure. 
\item \gls{sl} does not implicitly learn augmentation invariance, but learning to become invariant to SimCLR's augmentations implicitly fits the class structure and induces linear separability.
\item The representational structures rapidly diverge in the final layers, suggesting that SimCLR's performance stems from class-informative \textit{intermediate} representations, rather than implicit structural agreement between learned solutions to the \gls{sl} and SimCLR objectives. 
\end{itemize}

\begin{figure}[t]
    \centering
    \includegraphics[width=0.9\textwidth]{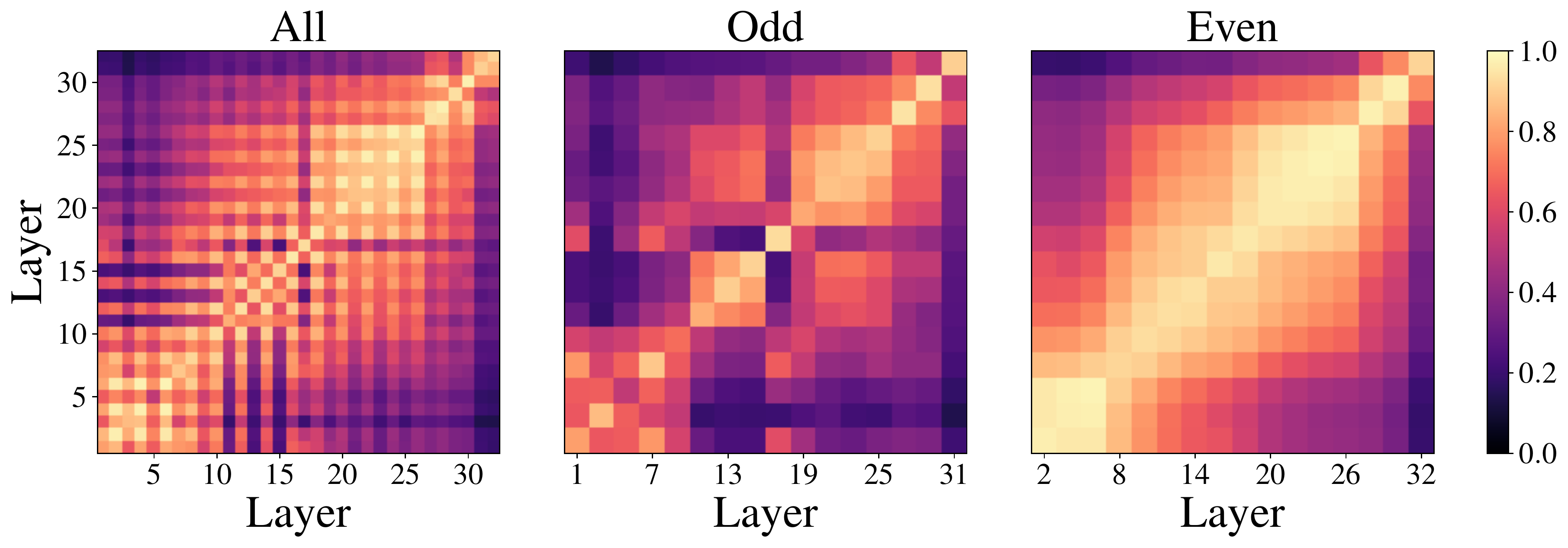}
    \caption{CKA between layers of R50 networks trained via SimCLR. We show all, odd, and even layers in the left, middle, and right plots respectively. 
    In contrast to prior work, we compare across different initializations as a sanity check for solution stability.}	
    \label{fig:internal_structure}
    \label{fig:simclr_vs_simclr}
\end{figure}

\setcounter{footnote}{0}

\section{Background}

\paragraph{Multi-view visual SSL} A number of recent \gls{ssl} algorithms for visual data focus on a view invariance auxiliary objective, where the model learns to identify different views of the same input image and distinguish views of different images. 
Here, we focus on SimCLR \citep{DBLP:conf/icml/ChenK0H20} as a step towards understanding contrastive self-supervised representation learning. We leave the analysis of alternative visual SSL methods for future work. 

\paragraph{SimCLR}
SimCLR learns representations by contrasting different views 
$\vv_t(\vx),\vv_{t^\prime}(\vx)$
of a single image $\vx$
to views
$\vv_{t^{\prime\prime}}(\vx_-),\vv_{t^{\prime\prime\prime}}(\vx_-)$
of other images, where we sample from a family of augmentations $t,t^\prime,t^{\prime\prime},t^{\prime\prime\prime}\sim\mathcal{T}$.
Views are constructed through  application: $\vv_t(\vx)=(g_{\vtheta} \circ f_{\vtheta} \circ t) (\vx)$,
where $f_{\vtheta}$ is the parametric backbone, typically a CNN, and 
$g_{\vtheta}$ is the \gls{nce} head, typically an MLP.
SimCLR's objective is then to minimize InfoNCE \citep{DBLP:conf/icml/ChenK0H20, DBLP:journals/corr/abs-1807-03748}:
\begin{align}
  \mathcal{L}_\text{InfoNCE}^{(i,j)}
  &= - \log\frac{\exp(\textnormal{sim}({\vv}_i, {\vv}_j) / \tau)}{\sum_{k=1}^{2N} {1}_{[k \neq i]} \exp(\textnormal{sim}({\vv}_i, {\vv}_k) / \tau)},
\end{align}
where 
$\vv_i=\vv_t(\vx)$, $\vv_j=\vv_{t^\prime}(\vx)$ 
are different views of the same image, 
$\vv_k=\vv_{t^{\prime\prime}}(\vx^{(k)}_-)$ 
are different views of different images, 
$\tau$ is the temperature,
and 
$\textnormal{sim}(\vu,\vv)=\vu\tran\vv/\pnorm\vu2\pnorm\vv2$ is cosine similarity.

\newcommand{\DBackbone}{D_{\textnormal{Backbone}}}

\paragraph{Comparing neural representation spaces}
Comparing neural representations is challenging due to their distributed nature, potential misalignment, and high dimensionality.  Prior work has demonstrated the utility of \gls{cka} as a similarity index which elegantly addresses these challenges \citep{DBLP:conf/icml/Kornblith0LH19}, enabling the analysis of a variety of neural architectures \citep{DBLP:conf/icml/Kornblith0LH19,DBLP:conf/iclr/NguyenRK21,DBLP:journals/corr/abs-2108-08810}. 

Let $\mX\in\R^{m\times p_1}$, $\mY\in\R^{m\times p_2}$ be $p_1$ and $p_2$ dimensional representation matrices whose rows are aligned\footnote{The $i$th row in $\mX$ and $\mY$ correspond to the $i$th sample for all $i\in1,\ldots,m$.}. 
Let $\mK=\mX\mX\tran$, $\mL=\mY\mY\tran$ be the corresponding Gram matrices. The \gls{cka} value is the normalized \gls{hsic} \citep{NIPS2007_d5cfead9} of these Gram matrices:
\begin{align}
    \cka(\mK,\mL)
    =
    \frac{
    \hsic(\mK,\mL)
    }{
    \sqrt{\hsic(\mK,\mK)\hsic(\mL,\mL)}
    },
\end{align}
We use the linear kernel due to its strong empirical performance and computational efficiency, simplifying the calculation of \gls{hsic} to 
$(m^2-1)\hsic(\mK,\mL)=\Tr(\mK\mH\mL\mH)=\vecc(\mK^\prime)\cdot \vecc(\mL^\prime)$, where $\mK^\prime=\mH\mK\mH$, $\mL^\prime=\mH\mL\mH$, and $\mH=\mI_m-\tfrac1m\vone\vone\tran$ is the centering matrix.

\paragraph{Experimental setup}
We use a R50
\citep{DBLP:conf/cvpr/HeZRS16} 
backbone for each model. 
For SimCLR, we train as specified in \citet{DBLP:conf/icml/ChenK0H20}. 
We group representations into residual (odd) and post-residual (even) layers, in line with the analysis of \citet{DBLP:conf/icml/Kornblith0LH19}. 
Further details are outlined in \Cref{app:experimental_setup}.

\newpage

\section{Results}

\begin{figure}[t]
    \centering
    \includegraphics[width=0.96\textwidth]{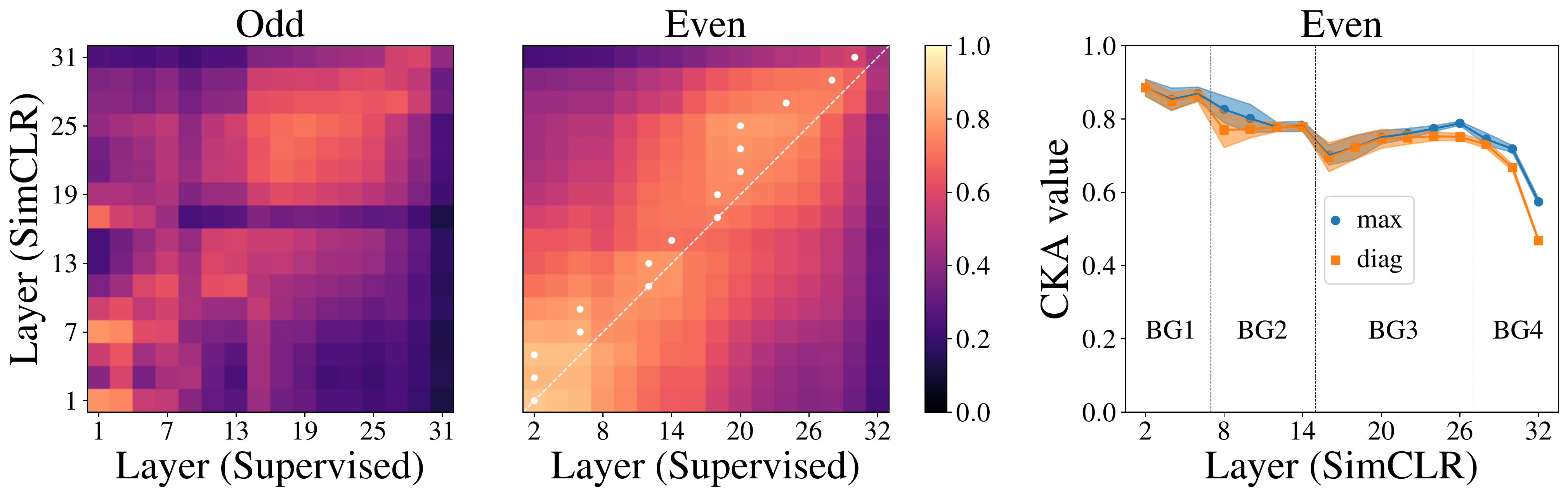}
    \caption{ (Left/Middle) 
    CKA between the odd/even layers of networks trained by SimCLR and \gls{sl}. For the evens, we mark the most similar \gls{sl} layer for each SimCLR layer with a white dot. 
    (Right) For each even layer in SimCLR, the similarity to its corresponding supervised layer (diag), and to the most similar supervised layer (max). We also denote the \gls{bg} (see \Cref{app:even-odd}).}
    \label{fig:simclr_v_supervised}
\end{figure}

\subsection{Internal representational structure of SimCLR}	

We begin by using \gls{cka} to study the internal representational similarity of SimCLR in Figure 1. This result mirrors the supervised analysis of \citet{DBLP:conf/icml/Kornblith0LH19},
indicating that \gls{sl} and SimCLR utilize the residual architecture in a similar way, with residual blocks decoupling from each other.
For completeness, we replicate the \gls{sl} result under our experimental setup in \Cref{app:supervised_internal_structure}.

\subsection{Comparing early and intermediate SimCLR and supervised representations} Next, we compare the representational structures induced by SimCLR and \gls{sl}. In
\Cref{fig:simclr_v_supervised}, we plot the odd and even layer CKA matrices across the learning methods, we observe:

\paragraph{Common primitives}
Residual representations are similar in the very early layers, perhaps due to both objectives inducing common primitives like Gabor filters \citep{DBLP:journals/jmlr/VincentLLBM10}.

\paragraph{Dissimilar residual (Odd)}
Beyond these initial layers, similarity between the residual representations substantially reduces,
indicating that each method learns residuals that operate on the input in qualitatively different ways -- likely a reflection of their distinct learning objectives. 

\paragraph{Similar post-residual (Even)}
Despite the dissimilarity of residuals, there is high similarity across the diagonal in the post-residual layers, indicating that the representations accumulated remain similar across learning methods;
similar representations are learned in a dissimilar way.

\paragraph{Stalling behaviour}
Finally, SimCLR appears to ``stall'' upon entering a new BG, remaining more similar to previous supervised layers, before ``catching up'' to the diagonal. 
This may be induced by SimCLR's strong augmentations, requiring a broader distribution to be compressed after each BG.

\begin{figure}[t]
    \centering
    \includegraphics[width=0.9\textwidth]{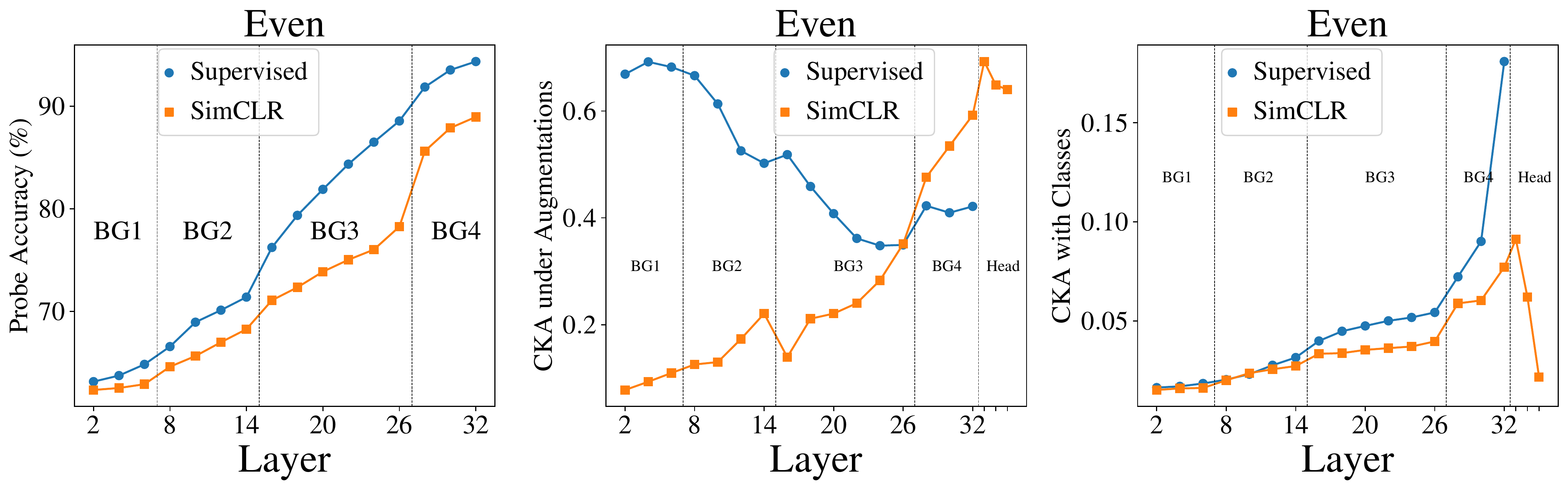}
    \caption{(Left) Linear probe accuracies for learned representations in the SimCLR and \gls{sl} models. (Middle) CKA between representations of differently augmented datasets at corresponding layers. (Right) CKA of learned representations with the class representations. We plot post-residual (even) layers only, denoting the block groups (BG) and NCE head (Head) where appropriate.}
    \label{fig:divergence}
\end{figure}

\subsection{Late layer representational dissimilarity of SimCLR and supervised learning}

\Cref{fig:simclr_v_supervised} (right) indicates that the representational structures learned by SimCLR and \gls{sl} rapidly diverge in the final block group. Here, we analyze this behaviour.

\paragraph{Linear separability of classes} 
We first investigate the effect that this divergence has on performance. We compute the accuracy of linear probes fitted at each layer in the SimCLR and supervised networks (\Cref{fig:divergence} (left)).
We find a monotonic increase in linear separability of the classes for both methods. This suggests that despite the divergence in later layers, both representational structures continue to become more linearly separable with respect to the classes. This raises the question: if the structures are diverging, but both are becoming more separable, what exactly is being learned?

\paragraph{Augmentation invariance} 
In \Cref{fig:divergence} (middle), we inspect what happens in the layers of both networks with respect to SimCLR's augmentation invariance objective. 
Here, we augment each sample in the C10 test dataset with two augmentations sampled from the augmentation distribution used during training\footnote{ImageNet augmentations for \gls{sl}, SimCLR augmentations for SimCLR.}, creating pairs of augmented test datasets. 
We measure the degree of invariance at each layer by plotting the CKA value between the representations of these augmented datasets. 
We observe that SimCLR's representations become more augmentation invariant with depth, increasingly so in the final few layers of the network. 
This contrasts with SL where the representations start out similar under (weaker) augmentations, then diverge until the final block group, where we see a small increase in CKA -- presumably due to classification. This result tells us (1) SimCLR does learn substantial augmentation invariance and (2) \gls{sl} does not implicitly learn augmentation invariant representations. This is perhaps surprising from the perspective of classification as a form of augmentation invariance where the augmentation distribution is the class-conditioned data distribution. Full CKA heatmaps are presented in \Cref{app:full_augmentation_invariance}.

\paragraph{Mapping to the classes}  Next we look at the \gls{sl} objective which, from a representation learning perspective, maps inputs to their assigned vertices on the simplex in the class representation space. In \Cref{fig:divergence} (right), we plot the CKA similarity between the class representations and the learned representations in the layers of the SimCLR and supervised networks. We observe a monotonic increase in CKA with the class structure for both methods throughout the backbone, offering insight into the increasing linear separability. It is however clear that \gls{sl}  accelerates much more rapidly towards the class structure in the final block group due to explicit optimization -- likely explaining the divergence of SL and SimCLR in \Cref{fig:simclr_v_supervised}. We also observe a decrease in similarity to the class structure after the first layer of the NCE head, perhaps revealing its role as a buffer which allows the backbone to learn richer class-informative features rather than immediately fit to InfoNCE.

\section{Conclusion}
We have shown the utility of CKA for comparing across \textit{learning methods}, rather than architectures. Using this approach, we have demonstrated that SimCLR representations are similar to those of supervised learning in their intermediate layers. Interestingly, we see divergence in the final few layers where each methods fits to its own objectives. 
Here, SimCLR learns augmentation invariance, contrasting with supervised learning, which instead is more strongly drawn to align with the labelled class structure.
This suggests that it is not similarity of the final representational structures that facilitates SimCLR's strong downstream performance. Rather, it is the similarity of the intermediate representations, i.e. the class-informative features that are learned along the way. 

These findings raise important questions for auxiliary task design: Can we build label-free tasks that share more intermediate features with supervised learning? Should we include inductive biases that look like ``mapping to the simplex'', e.g. orthogonality? Is mapping to the simplex desirable? Or are self-supervised representations more robust in a multi-task/multi-distribution setting? We leave these questions for future work.

\section*{Acknowledgements}
The authors would like to thank the following people for their help throughout the process of writing this paper, in alphabetical order:
Barry-John Theobald,
Luca Zappella, 
and Xavier Suau Cuadros.
Additionally,
we thank 
Andrea Klein,
Cindy Liu,
Guihao Liang,
Guillaume Seguin,
Li Li,
Okan Akalin,
and the wider
Apple infrastructure team for assistance with developing scalable, fault tolerant
code.

\bibliography{
libraries/distributions,
libraries/selfsup,
libraries/celeba,
libraries/models,
libraries/other,
libraries/cka.bib,
libraries/cnn.bib
}

\begin{thebibliography}{24}
\providecommand{\natexlab}[1]{#1}
\providecommand{\url}[1]{\texttt{#1}}
\expandafter\ifx\csname urlstyle\endcsname\relax
  \providecommand{\doi}[1]{doi: #1}\else
  \providecommand{\doi}{doi: \begingroup \urlstyle{rm}\Url}\fi

\bibitem[Deng et~al.(2009)Deng, Dong, Socher, Li, Li, and
  Li]{DBLP:conf/cvpr/DengDSLL009}
Jia Deng, Wei Dong, Richard Socher, Li{-}Jia Li, Kai Li, and Fei{-}Fei Li.
\newblock Imagenet: {A} large-scale hierarchical image database.
\newblock In \emph{2009 {IEEE} Computer Society Conference on Computer Vision
  and Pattern Recognition {(CVPR} 2009), 20-25 June 2009, Miami, Florida,
  {USA}}, pages 248--255. {IEEE} Computer Society, 2009.
\newblock \doi{10.1109/CVPR.2009.5206848}.
\newblock URL \url{https://doi.org/10.1109/CVPR.2009.5206848}.

\bibitem[Dosovitskiy et~al.(2021)Dosovitskiy, Beyer, Kolesnikov, Weissenborn,
  Zhai, Unterthiner, Dehghani, Minderer, Heigold, Gelly, Uszkoreit, and
  Houlsby]{DBLP:conf/iclr/DosovitskiyB0WZ21}
Alexey Dosovitskiy, Lucas Beyer, Alexander Kolesnikov, Dirk Weissenborn,
  Xiaohua Zhai, Thomas Unterthiner, Mostafa Dehghani, Matthias Minderer, Georg
  Heigold, Sylvain Gelly, Jakob Uszkoreit, and Neil Houlsby.
\newblock An image is worth 16x16 words: Transformers for image recognition at
  scale.
\newblock In \emph{9th International Conference on Learning Representations,
  {ICLR} 2021, Virtual Event, Austria, May 3-7, 2021}. OpenReview.net, 2021.
\newblock URL \url{https://openreview.net/forum?id=YicbFdNTTy}.

\bibitem[He et~al.(2016)He, Zhang, Ren, and Sun]{DBLP:conf/cvpr/HeZRS16}
Kaiming He, Xiangyu Zhang, Shaoqing Ren, and Jian Sun.
\newblock Deep residual learning for image recognition.
\newblock In \emph{2016 {IEEE} Conference on Computer Vision and Pattern
  Recognition, {CVPR} 2016, Las Vegas, NV, USA, June 27-30, 2016}, pages
  770--778. {IEEE} Computer Society, 2016.
\newblock \doi{10.1109/CVPR.2016.90}.
\newblock URL \url{https://doi.org/10.1109/CVPR.2016.90}.

\bibitem[Krizhevsky et~al.(2012)Krizhevsky, Sutskever, and
  Hinton]{DBLP:conf/nips/KrizhevskySH12}
Alex Krizhevsky, Ilya Sutskever, and Geoffrey~E. Hinton.
\newblock Imagenet classification with deep convolutional neural networks.
\newblock In Peter~L. Bartlett, Fernando C.~N. Pereira, Christopher J.~C.
  Burges, L{\'{e}}on Bottou, and Kilian~Q. Weinberger, editors, \emph{Advances
  in Neural Information Processing Systems 25: 26th Annual Conference on Neural
  Information Processing Systems 2012. Proceedings of a meeting held December
  3-6, 2012, Lake Tahoe, Nevada, United States}, pages 1106--1114, 2012.
\newblock URL
  \url{https://proceedings.neurips.cc/paper/2012/hash/c399862d3b9d6b76c8436e924a68c45b-Abstract.html}.

\bibitem[LeCun et~al.(1989)LeCun, Boser, Denker, Henderson, Howard, Hubbard,
  and Jackel]{DBLP:journals/neco/LeCunBDHHHJ89}
Yann LeCun, Bernhard~E. Boser, John~S. Denker, Donnie Henderson, Richard~E.
  Howard, Wayne~E. Hubbard, and Lawrence~D. Jackel.
\newblock Backpropagation applied to handwritten zip code recognition.
\newblock \emph{Neural Comput.}, 1\penalty0 (4):\penalty0 541--551, 1989.
\newblock \doi{10.1162/neco.1989.1.4.541}.
\newblock URL \url{https://doi.org/10.1162/neco.1989.1.4.541}.

\bibitem[Russakovsky et~al.(2014)Russakovsky, Deng, Su, Krause, Satheesh, Ma,
  Huang, Karpathy, Khosla, Bernstein, Berg, and
  Fei{-}Fei]{DBLP:journals/corr/RussakovskyDSKSMHKKBBF14}
Olga Russakovsky, Jia Deng, Hao Su, Jonathan Krause, Sanjeev Satheesh, Sean Ma,
  Zhiheng Huang, Andrej Karpathy, Aditya Khosla, Michael~S. Bernstein,
  Alexander~C. Berg, and Li~Fei{-}Fei.
\newblock Imagenet large scale visual recognition challenge.
\newblock \emph{CoRR}, abs/1409.0575, 2014.
\newblock URL \url{http://arxiv.org/abs/1409.0575}.

\bibitem[Zeiler and Fergus(2014)]{DBLP:conf/eccv/ZeilerF14}
Matthew~D. Zeiler and Rob Fergus.
\newblock Visualizing and understanding convolutional networks.
\newblock In David~J. Fleet, Tom{\'{a}}s Pajdla, Bernt Schiele, and Tinne
  Tuytelaars, editors, \emph{Computer Vision - {ECCV} 2014 - 13th European
  Conference, Zurich, Switzerland, September 6-12, 2014, Proceedings, Part
  {I}}, volume 8689 of \emph{Lecture Notes in Computer Science}, pages
  818--833. Springer, 2014.
\newblock \doi{10.1007/978-3-319-10590-1\_53}.
\newblock URL \url{https://doi.org/10.1007/978-3-319-10590-1\_53}.

\bibitem[Alayrac et~al.(2020)Alayrac, Recasens, Schneider, Arandjelovic,
  Ramapuram, Fauw, Smaira, Dieleman, and
  Zisserman]{DBLP:conf/nips/AlayracRSARFSDZ20}
Jean{-}Baptiste Alayrac, Adri{\`{a}} Recasens, Rosalia Schneider, Relja
  Arandjelovic, Jason Ramapuram, Jeffrey~De Fauw, Lucas Smaira, Sander
  Dieleman, and Andrew Zisserman.
\newblock Self-supervised multimodal versatile networks.
\newblock In Hugo Larochelle, Marc'Aurelio Ranzato, Raia Hadsell,
  Maria{-}Florina Balcan, and Hsuan{-}Tien Lin, editors, \emph{Advances in
  Neural Information Processing Systems 33: Annual Conference on Neural
  Information Processing Systems 2020, NeurIPS 2020, December 6-12, 2020,
  virtual}, 2020.
\newblock URL
  \url{https://proceedings.neurips.cc/paper/2020/hash/0060ef47b12160b9198302ebdb144dcf-Abstract.html}.

\bibitem[Caron et~al.(2020)Caron, Misra, Mairal, Goyal, Bojanowski, and
  Joulin]{DBLP:conf/nips/CaronMMGBJ20}
Mathilde Caron, Ishan Misra, Julien Mairal, Priya Goyal, Piotr Bojanowski, and
  Armand Joulin.
\newblock Unsupervised learning of visual features by contrasting cluster
  assignments.
\newblock In Hugo Larochelle, Marc'Aurelio Ranzato, Raia Hadsell,
  Maria{-}Florina Balcan, and Hsuan{-}Tien Lin, editors, \emph{Advances in
  Neural Information Processing Systems 33: Annual Conference on Neural
  Information Processing Systems 2020, NeurIPS 2020, December 6-12, 2020,
  virtual}, 2020.
\newblock URL
  \url{https://proceedings.neurips.cc/paper/2020/hash/70feb62b69f16e0238f741fab228fec2-Abstract.html}.

\bibitem[Caron et~al.(2021)Caron, Touvron, Misra, J{\'{e}}gou, Mairal,
  Bojanowski, and Joulin]{DBLP:journals/corr/abs-2104-14294}
Mathilde Caron, Hugo Touvron, Ishan Misra, Herv{\'{e}} J{\'{e}}gou, Julien
  Mairal, Piotr Bojanowski, and Armand Joulin.
\newblock Emerging properties in self-supervised vision transformers.
\newblock \emph{CoRR}, abs/2104.14294, 2021.
\newblock URL \url{https://arxiv.org/abs/2104.14294}.

\bibitem[Chen et~al.(2020{\natexlab{a}})Chen, Kornblith, Norouzi, and
  Hinton]{DBLP:conf/icml/ChenK0H20}
Ting Chen, Simon Kornblith, Mohammad Norouzi, and Geoffrey~E. Hinton.
\newblock A simple framework for contrastive learning of visual
  representations.
\newblock In \emph{Proceedings of the 37th International Conference on Machine
  Learning, {ICML} 2020, 13-18 July 2020, Virtual Event}, volume 119 of
  \emph{Proceedings of Machine Learning Research}, pages 1597--1607. {PMLR},
  2020{\natexlab{a}}.
\newblock URL \url{http://proceedings.mlr.press/v119/chen20j.html}.

\bibitem[Chen et~al.(2020{\natexlab{b}})Chen, Kornblith, Swersky, Norouzi, and
  Hinton]{DBLP:conf/nips/ChenKSNH20}
Ting Chen, Simon Kornblith, Kevin Swersky, Mohammad Norouzi, and Geoffrey~E.
  Hinton.
\newblock Big self-supervised models are strong semi-supervised learners.
\newblock In Hugo Larochelle, Marc'Aurelio Ranzato, Raia Hadsell,
  Maria{-}Florina Balcan, and Hsuan{-}Tien Lin, editors, \emph{Advances in
  Neural Information Processing Systems 33: Annual Conference on Neural
  Information Processing Systems 2020, NeurIPS 2020, December 6-12, 2020,
  virtual}, 2020{\natexlab{b}}.
\newblock URL
  \url{https://proceedings.neurips.cc/paper/2020/hash/fcbc95ccdd551da181207c0c1400c655-Abstract.html}.

\bibitem[Grill et~al.(2020)Grill, Strub, Altch{\'{e}}, Tallec, Richemond,
  Buchatskaya, Doersch, Pires, Guo, Azar, Piot, Kavukcuoglu, Munos, and
  Valko]{DBLP:conf/nips/GrillSATRBDPGAP20}
Jean{-}Bastien Grill, Florian Strub, Florent Altch{\'{e}}, Corentin Tallec,
  Pierre~H. Richemond, Elena Buchatskaya, Carl Doersch, Bernardo~{\'{A}}vila
  Pires, Zhaohan Guo, Mohammad~Gheshlaghi Azar, Bilal Piot, Koray Kavukcuoglu,
  R{\'{e}}mi Munos, and Michal Valko.
\newblock Bootstrap your own latent - {A} new approach to self-supervised
  learning.
\newblock In Hugo Larochelle, Marc'Aurelio Ranzato, Raia Hadsell,
  Maria{-}Florina Balcan, and Hsuan{-}Tien Lin, editors, \emph{Advances in
  Neural Information Processing Systems 33: Annual Conference on Neural
  Information Processing Systems 2020, NeurIPS 2020, December 6-12, 2020,
  virtual}, 2020.
\newblock URL
  \url{https://proceedings.neurips.cc/paper/2020/hash/f3ada80d5c4ee70142b17b8192b2958e-Abstract.html}.

\bibitem[Zbontar et~al.(2021)Zbontar, Jing, Misra, LeCun, and
  Deny]{DBLP:journals/corr/abs-2103-03230}
Jure Zbontar, Li~Jing, Ishan Misra, Yann LeCun, and St{\'{e}}phane Deny.
\newblock Barlow twins: Self-supervised learning via redundancy reduction.
\newblock \emph{CoRR}, abs/2103.03230, 2021.
\newblock URL \url{https://arxiv.org/abs/2103.03230}.

\bibitem[van~den Oord et~al.(2018)van~den Oord, Li, and
  Vinyals]{DBLP:journals/corr/abs-1807-03748}
A{\"{a}}ron van~den Oord, Yazhe Li, and Oriol Vinyals.
\newblock Representation learning with contrastive predictive coding.
\newblock \emph{CoRR}, abs/1807.03748, 2018.
\newblock URL \url{http://arxiv.org/abs/1807.03748}.

\bibitem[Kornblith et~al.(2019)Kornblith, Norouzi, Lee, and
  Hinton]{DBLP:conf/icml/Kornblith0LH19}
Simon Kornblith, Mohammad Norouzi, Honglak Lee, and Geoffrey~E. Hinton.
\newblock Similarity of neural network representations revisited.
\newblock In Kamalika Chaudhuri and Ruslan Salakhutdinov, editors,
  \emph{Proceedings of the 36th International Conference on Machine Learning,
  {ICML} 2019, 9-15 June 2019, Long Beach, California, {USA}}, volume~97 of
  \emph{Proceedings of Machine Learning Research}, pages 3519--3529. {PMLR},
  2019.
\newblock URL \url{http://proceedings.mlr.press/v97/kornblith19a.html}.

\bibitem[Nguyen et~al.(2021)Nguyen, Raghu, and
  Kornblith]{DBLP:conf/iclr/NguyenRK21}
Thao Nguyen, Maithra Raghu, and Simon Kornblith.
\newblock Do wide and deep networks learn the same things? uncovering how
  neural network representations vary with width and depth.
\newblock In \emph{9th International Conference on Learning Representations,
  {ICLR} 2021, Virtual Event, Austria, May 3-7, 2021}. OpenReview.net, 2021.
\newblock URL \url{https://openreview.net/forum?id=KJNcAkY8tY4}.

\bibitem[Raghu et~al.(2021)Raghu, Unterthiner, Kornblith, Zhang, and
  Dosovitskiy]{DBLP:journals/corr/abs-2108-08810}
Maithra Raghu, Thomas Unterthiner, Simon Kornblith, Chiyuan Zhang, and Alexey
  Dosovitskiy.
\newblock Do vision transformers see like convolutional neural networks?
\newblock \emph{CoRR}, abs/2108.08810, 2021.
\newblock URL \url{https://arxiv.org/abs/2108.08810}.

\bibitem[Gretton et~al.(2008)Gretton, Fukumizu, Teo, Song, Sch\"{o}lkopf, and
  Smola]{NIPS2007_d5cfead9}
Arthur Gretton, Kenji Fukumizu, Choon Teo, Le~Song, Bernhard Sch\"{o}lkopf, and
  Alex Smola.
\newblock A kernel statistical test of independence.
\newblock In J.~Platt, D.~Koller, Y.~Singer, and S.~Roweis, editors,
  \emph{Advances in Neural Information Processing Systems}, volume~20. Curran
  Associates, Inc., 2008.
\newblock URL
  \url{https://proceedings.neurips.cc/paper/2007/file/d5cfead94f5350c12c322b5b664544c1-Paper.pdf}.

\bibitem[Vincent et~al.(2010)Vincent, Larochelle, Lajoie, Bengio, and
  Manzagol]{DBLP:journals/jmlr/VincentLLBM10}
Pascal Vincent, Hugo Larochelle, Isabelle Lajoie, Yoshua Bengio, and
  Pierre{-}Antoine Manzagol.
\newblock Stacked denoising autoencoders: Learning useful representations in a
  deep network with a local denoising criterion.
\newblock \emph{J. Mach. Learn. Res.}, 11:\penalty0 3371--3408, 2010.
\newblock URL \url{http://portal.acm.org/citation.cfm?id=1953039}.

\bibitem[Huo et~al.(2021)Huo, Gu, and Huang]{DBLP:conf/aaai/HuoGH21}
Zhouyuan Huo, Bin Gu, and Heng Huang.
\newblock Large batch optimization for deep learning using new complete
  layer-wise adaptive rate scaling.
\newblock In \emph{Thirty-Fifth {AAAI} Conference on Artificial Intelligence,
  {AAAI} 2021, Thirty-Third Conference on Innovative Applications of Artificial
  Intelligence, {IAAI} 2021, The Eleventh Symposium on Educational Advances in
  Artificial Intelligence, {EAAI} 2021, Virtual Event, February 2-9, 2021},
  pages 7883--7890. {AAAI} Press, 2021.
\newblock URL \url{https://ojs.aaai.org/index.php/AAAI/article/view/16962}.

\bibitem[Goyal et~al.(2017)Goyal, Doll{\'{a}}r, Girshick, Noordhuis,
  Wesolowski, Kyrola, Tulloch, Jia, and He]{DBLP:journals/corr/GoyalDGNWKTJH17}
Priya Goyal, Piotr Doll{\'{a}}r, Ross~B. Girshick, Pieter Noordhuis, Lukasz
  Wesolowski, Aapo Kyrola, Andrew Tulloch, Yangqing Jia, and Kaiming He.
\newblock Accurate, large minibatch {SGD:} training imagenet in 1 hour.
\newblock \emph{CoRR}, abs/1706.02677, 2017.
\newblock URL \url{http://arxiv.org/abs/1706.02677}.

\bibitem[Smith and Topin(2017)]{DBLP:journals/corr/abs-1708-07120}
Leslie~N. Smith and Nicholay Topin.
\newblock Super-convergence: Very fast training of residual networks using
  large learning rates.
\newblock \emph{CoRR}, abs/1708.07120, 2017.
\newblock URL \url{http://arxiv.org/abs/1708.07120}.

\bibitem[Krizhevsky and Hinton(2009)]{Krizhevsky_2009_17719}
Alex Krizhevsky and Geoffrey Hinton.
\newblock Learning multiple layers of features from tiny images.
\newblock Technical Report~0, University of Toronto, Toronto, Ontario, 2009.

\end{thebibliography}

\newpage

\appendix

\section{Experimental Setup}
\label{app:experimental_setup}

\paragraph{Experimental setup}

We choose the same ResNet50 architecture
\citep{DBLP:conf/cvpr/HeZRS16} 
for SimCLR and the supervised model.
We follow the training procedure described in \citet{DBLP:conf/icml/ChenK0H20}:
all models use the LARS optimizer
\citep{DBLP:conf/aaai/HuoGH21} with linear warmup
\citep{DBLP:journals/corr/GoyalDGNWKTJH17} and a single cycle cosine annealed
learning rate schedule
\citep{DBLP:journals/corr/GoyalDGNWKTJH17,DBLP:journals/corr/abs-1708-07120}. 
SimCLR models are trained for 1300 epochs with a batch size of 4096 under SimCLR augmentations \citep{DBLP:conf/icml/ChenK0H20}, 
whereas our supervised models are trained for 300 epochs using a batch size of 8192 under standard ImageNet augmentations\footnote{
\texttt{RandomResizedCrop(224)}, \texttt{RandomHorizontalFlip} and channel-wise standardisation.}. For SimCLR, we implement the original version of \citet{DBLP:conf/icml/ChenK0H20}, where in particular the NCE head is a 3-layer MLP.

For each learning method, we train from 3 different random initializations, resulting in 6 models.
Models are trained on the training set of CIFAR-10 (50,000 samples) \citep{Krizhevsky_2009_17719}.
Representations are produced on the test set of CIFAR-10 (10,000 samples) under each model's corresponding test augmentation family.
Representations are flattened, producing a single vector for each sample.

\subsection{Even and Odd representations}
\label{app:even-odd}

For each bottleneck layer, we extract the following representations:
\begin{python}
class Bottleneck(nn.Module):
    # Other definitions

    def forward(self, x: Tensor) -> Tensor:
        identity = x

        out = self.conv1(x)
        out = self.bn1(out)
        out = self.relu(out)

        out = self.conv2(out)
        out = self.bn2(out)
        out = self.relu(out)

        out = self.conv3(out)
        ODD_REPRESENTATIONS[i] = out = self.bn3(out)

        if self.downsample is not None:
            identity = self.downsample(x)

        out += identity.       
        EVEN_REPRESENTATIONS[i] = out = self.relu(out)

        return out
\end{python}
i.e. two representations per bottleneck.

A ResNet50 is built out of 4 block groups, each subsequent group increasing dimensionality (see \Cref{tab:bg-size}).
The total number of bottleneck layers across all groups is $16=3+4+6+3$, resulting in 16 odd representations and 16 even representations that we use in our analysis.

\begin{table}[ht]
    \centering
    \caption{The filter properties and bottleneck multiplicities in each \gls{bg} of a ResNet50.}
    \label{tab:bg-size}    
    \begin{tabular}{ccc}
    \toprule
         Group Name & Number of Bottlenecks & Filters in each Bottleneck \\ \midrule
         BG1 & 3 & $(1\times1,64)$, $(3\times3,64)$, $(1\times1,256)$ \\
         BG2 & 4 & $(1\times1,128)$, $(3\times3,128)$, $(1\times1,512)$ \\
         BG3 & 6 & $(1\times1,256)$, $(3\times3,256)$, $(1\times1,1024)$ \\
         BG4 & 3 & $(1\times1,512)$, $(3\times3,512)$, $(1\times1,2048)$ \\ \bottomrule
    \end{tabular}
\end{table}

\newpage

\newpage 
\section{Internal representational structure for supervised learning}
\label{app:supervised_internal_structure}

\begin{figure}[t]
    \centering
    \includegraphics[width=0.9\textwidth]{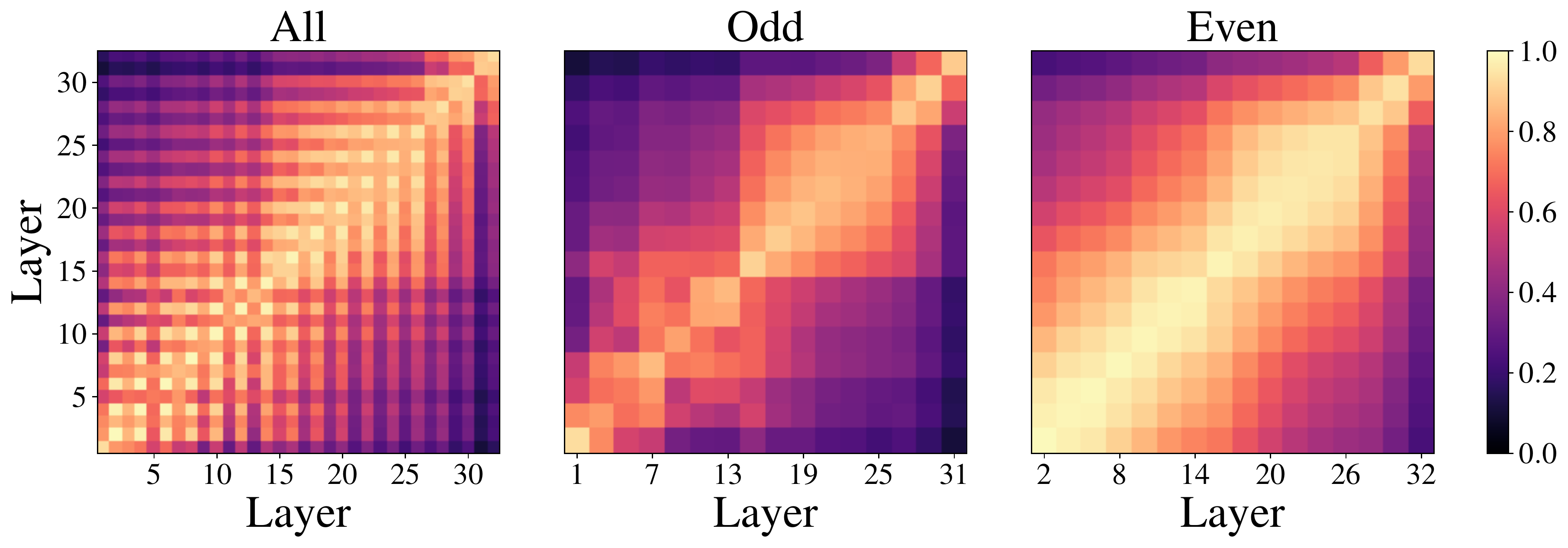}
    \caption{\gls{cka} between all layers of ResNet-50 networks trained via supervision. We plot all layers in the left column, and even/odd layers on the middle/right.}
    \label{fig:supervised_internal}
\end{figure}

Here, we replicate the results of \citet{DBLP:conf/icml/Kornblith0LH19} in our experimental setup. In particular, in \Cref{fig:supervised_internal} we use CKA to compare the learned representations of ResNet-50 architectures trained via supervised learning, as specified in \Cref{app:experimental_setup}. We note that in contrast to their work, we compare across different initializations in order to check for solution stability.

Corroborating their findings, we observe high similarity across neighbouring post-residual (even) layers in the network, and greater dissimilarity between residual (odd) layers, which largely appear similar only to themselves. The similarity of even layers is explained by the residual connections propagating representations through the network. The dissimilarity of odd layers suggests that each sequential block performs a distinct modification to this propagated residual representation. The similarity within block groups (i.e. at the same dimensionality) is higher than across block groups for all layers. We further note that the results for SimCLR (\Cref{fig:internal_structure}) mirrors that of supervised learning, except there appears to be even greater disagreement across block groups.

\section{Full Augmentation Invariance CKA Heatmaps}
\label{app:full_augmentation_invariance}

\begin{figure}[h]
    \centering
    \includegraphics[width=0.65\textwidth]{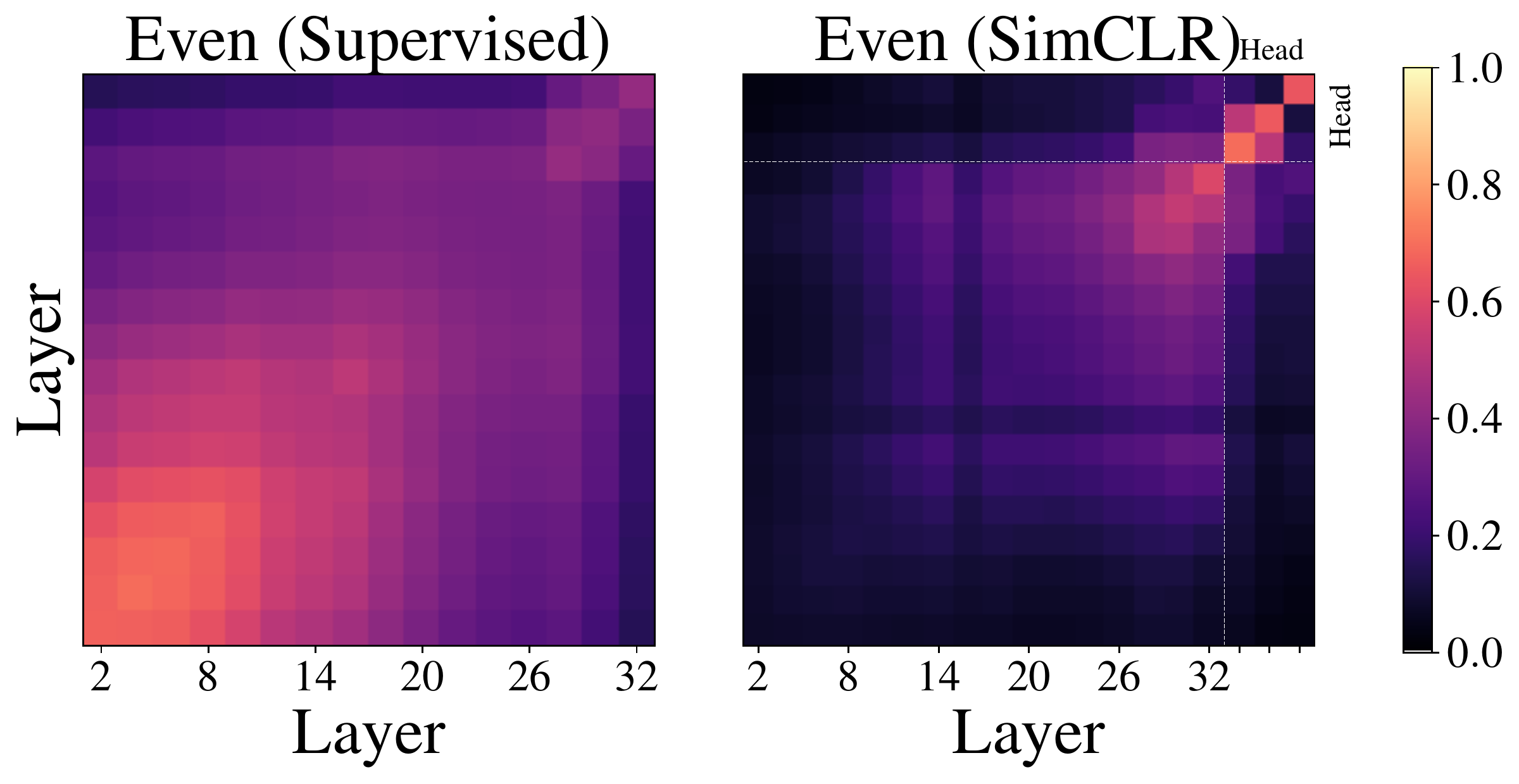}
    \caption{We apply the method-specific training augmentations to the CIFAR-10 test dataset and plot the CKA of the representations as they propagate through the supervised and SimCLR  models.}
    \label{fig:aug_invariance}
\end{figure}

In \Cref{fig:aug_invariance} we provide the all-layers CKA comparisons of the representations of differently augmented test datasets in the same model. The diagonals of \Cref{fig:aug_invariance} correspond to \Cref{fig:divergence} (middle). The augmentation invariance of the supervised model gradually fades starting from the bottom left corner, suggesting that it is initially due to the residual connections and weak augmentation strategy. The SimCLR plot is striking: substantial (but not total) invariance is learned in the NCE head, and this backpropagates into the final few layers of the backbone. However, the representations show limited robustness to augmentation right up until these last few layers of the network.

\end{document}